\documentclass{sn-jnl}

\usepackage{graphicx}%
\usepackage{multirow}%
\usepackage{amsmath,amssymb,amsfonts}%
\usepackage{amsthm}%
\usepackage{mathrsfs}%
\usepackage[title]{appendix}%
\usepackage{xcolor}%
\usepackage{textcomp}%
\usepackage{manyfoot}%
\usepackage{booktabs}%
\usepackage{algorithm}
\usepackage{tabularx}
\usepackage{natbib}
\usepackage{algorithmicx}%
\usepackage{algpseudocode}%
\usepackage{listings}%

\theoremstyle{thmstyleone}%
\theoremstyle{thmstylethree}%

\begin{document}

\title[Article Title]{Competitive Task Allocation with Temporal Proposal Windows and Improved Artificial Potential Fields for Dual-Arm Industrial Manipulation}


\author*[1]{\fnm{Surya Prakash} \sur{S.K}}\email{d21077@students.iitmandi.ac.in}

\author[2]{\fnm{Bhuvan} \sur{Narula}}\email{bhuvan.narula@addverb.com}

\author[4]{\fnm{Jagannath Prasad} \sur{Sahoo}}\email{s23118@students.iitmandi.ac.in}

\author[3]{\fnm{Jayakumar} \sur{VK}}\email{jayakumar.vandavasi@dewa.gov.ae}
\author[4]{\fnm{Praful} \sur{Hambarde}}\email{praful@iitmandi.ac.in}
\author*[4]{\fnm{Amit} \sur{Shukla}}\email{amitshukla@iitmandi.ac.in}


\affil*[1]{\orgdiv{School of Mechanical and Materials Engineering}, \orgname{IIT Mandi}, \orgaddress{\street{Kamand}, \city{Mandi}, \postcode{175005}, \state{H.P}, \country{India}}}

\affil[2]{\orgdiv{R\&D}, \orgname{Addverb Technologies Limited}, \orgaddress{\street{Sec-156, Phase-II}, \city{Noida}, \postcode{201310}, \state{UP}, \country{India}}}

\affil[3]{\orgdiv{R\&D 4IR}, \orgname{Dubai Electricity \& Water Authority}, \orgaddress{\street{}, \city{Dubai}, \postcode{P.O. Box 564}, \state{}, \country{UAE}}}
\affil*[4]{\orgdiv{Centre for Artificial Intelligence and Robotics}, \orgname{IIT Mandi}, \orgaddress{\street{Kamand}, \city{Mandi}, \postcode{175005}, \state{H.P}, \country{India}}}


\abstract{Modern industrial automation demands autonomous collaborative systems capable of coordinated manipulation in workspaces with heterogeneous object arrangements. Existing systems rely on predetermined trajectories and rigid allocation strategies, causing suboptimal workspace utilization and collision risks. This paper presents a distance-based competitive task allocation framework for dual-arm systems, integrating real-time vision perception with collision-free motion planning. A temporal window mechanism enables each arm to evaluate object detections over five consecutive vision frames before committing to optimal distance-based selection, featuring dynamic reallocation and spatial conflict resolution at 0.01m threshold accuracy. An improved Artificial Potential Field (iAPF) framework provides coordinated dual-arm movements through exponential goal attraction, inter-arm repulsion, home-seeking, and velocity damping forces. Simulation results validate superiority over market-based, round-robin, and fixed allocation strategies in distance efficiency, collision risk, and workspace coverage. Hardware experiments using two 6-DOF Kinova Gen3 Lite manipulators demonstrate $\approx$47\% average velocity improvement over fixed allocation in homogeneous scenarios with zero collisions, while mixed object scenarios confirm universal handling with $\approx$11\% time overhead offset by superior safety margins. The framework integrates YOLOv8 OBB detection at 30 FPS with 10 Hz control loops, demonstrating practical deployment capability for industrial assembly and sorting operations.}

\keywords{Intelligent Robots, Competitive Task Allocations, Industrial Automation, Motion Planning, Artificial Potential Fields.}



\maketitle

\section{Introduction}
\label{intro}
Robots play a key role in transforming industries into smart manufacturing systems. Traditional production methods rely on either a labor-intensive human workforce or rigidly programmed robots to complete manufacturing tasks. While humans excel at high-dexterity tasks and can adapt intelligently to dynamic scenarios, they are prone to fatigue and errors during continuous, long-term operations \cite{T-mech4}. Conversely, robots can operate over extended periods but lack human-like intelligence and adaptability. Combining these complementary capabilities within a unified system will significantly advance autonomy in manufacturing systems \cite{intro_1}. Industrial setups have evolved around human bi-manual capabilities, with ergonomic guidelines ensuring optimal reach zones and bilateral manipulation spaces that accommodate natural two-arm coordination patterns. Modern manufacturing increasingly demands dual-arm coordination for assembly tasks in shared workspaces \cite{qin2024online}. Automotive assembly operations require simultaneous bi-manual manipulation where traditional sequential operations with fixed allocation cause idle time and workspace under utilization \cite{marvel2018multi,vision_1}.

Efficient task allocation in robot systems enhances overall performance and resolves workspace reachability constraints \cite{intro_4, intro_11, patra2024}. Current industrial practice assigns fixed object types to each arm, failing when objects appear outside designated zones is a critical limitation in flexible manufacturing cells \cite{pedersen2016robot}. Our framework addresses this gap through competitive allocation with integrated spatial conflict resolution during task assignment, enabling adaptability for unstructured industrial environments. 

Collaborative robotic systems, especially dual-arm manipulators with natural two-arm capabilities, competitive dynamic task allocation, and collision-free coordination, represent a significant advancement in manufacturing automation. Such systems can seamlessly integrate with existing ergonomic industrial designs, while advancing autonomous manufacturing capabilities beyond the limitations of traditional single-arm or pre-programmed robotic solutions.

The present paper is organized as following, Sec.\ref{intro} introduces dual-arm systems significance in manufacturing automation and establishes research motivation for Competitive task allocation with collision avoidance. While, Sec.\ref{background}, reviews existing task allocation methods, collision detection approaches, and highlights the gap in current dual-arm coordination systems. Sec.\ref{method} explains, comprehensive methodology combining four components, continuous inter-arm link distance computation, vision-based object detection, Competitive bucket management with temporal proposal windows, and improved APF framework. Sec.\ref{results} demonstrates experimental validation using two Kinova Gen3 Lite manipulators. Sec.\ref{concl} concludes with key findings, practical implications, and outlines future research directions for enhanced industrial deployment.

\section{Related Work}
\label{background}
The challenges of Competitive task allocation and safe collision-free operation in dual-arm systems have motivated significant research across robotics and automation fields. Market-based task allocation~\cite{aziz2022,galati, quinton} methods have emerged as a prominent distributed approach, where robots bid for tasks based on their reachability and current states. \cite{intro_8} proposed a dynamic threshold-based system using adaptive incentive value (AIV) computation with sequential single-item auctions and MiniSum/MiniMax optimization. However, centralized AIV computation and continuous system-wide resource tracking create computational bottlenecks unsuitable for real-time applications. \cite{greedymarket, rinaldi2023multi} proposed a greedy coalition auction algorithm for spatially distributed multi-agent 
systems, enabling dynamic task reassignment as agents move toward targets through decentralized bid negotiations. Furthermore, round-robin task allocation provides computational simplicity and straight forward task allocations but fails to account for task complexity, and spatial conflicts handling, leading to suboptimal utilization for spatially constrained workspaces\cite{korsah2013comprehensive}\cite{schneider2015auction}.

Recent learning-based approaches employ deep reinforcement learning (DRL) for adaptive task allocation. \cite{ma2024end, lu2024reinforcement} have shown promise in multi-robot coordination, with DRL approaches achieving better performance over heuristic baselines through end-to-end policy optimization. These methods primarily address spatially distributed scenarios in which robots operate in large-scale workspaces such as warehouses and agricultural fields. These multi-robot systems learn task-scheduling decisions from abstract state representations, such as resource availability, geometric proximity, and task priorities. For instance, \cite{huang2025reinforcement} formulated task allocation as a centralized reinforcement learning problem utilizing action-selective double Q-learning. In this framework, tasks are scheduled based on the robots' current positions, predetermined target distances, and battery states. Conversely, \cite{gautier2020comparison} proposed a decentralized Deep Q-Network (DQN)-based approach where robots learn to make run, transfer, or postpone decisions. This learning process relies on local resource usage and task priority as observations, notably excluding physical task locations or multi-robot spatial relationships. Moreover these learning based methods require extensive offline training and intensive iterations before taking decision compared to the reactive methods.

In dual-arm systems, where task allocation presents unique challenges due to shared workspace constraints, kinematic coupling, and the need for coordinated bilateral manipulation in industrial environments. \cite{intro_7} employed a three-layered A*/anytime A* architecture with constraint-based conflict resolution and stepwise rescheduling for online disruptions. While achieving good performance in structured construction tasks, the multi-layer planning architecture and constraint resolution approach may not be optimal for high-frequency object detection scenarios requiring immediate proximity-based allocation decisions. \cite{intro_9} proposed hybrid motion and task planning using AND/OR manipulation task graphs with motion planner feedback. While providing systematic solutions through backtracking search, the approach requires complete workspace knowledge and pre-computation of grasp configurations, making it unsuitable for dynamic real-time environments. \cite{intro_10} addressed dual-arm coordination using coordination diagrams with batch and incremental methods for task assignment. Although computationally efficient, both approaches assume known object placements and require collision-free trajectory pre-computation, limiting applicability to vision-based dynamic scenarios.

Once task allocation is determined, the fundamental challenge shifts to coordinated motion execution while ensuring collision-free trajectories in shared workspace environments\cite{vahrenkamp2009humanoid}. The classical APF framework introduced by \cite{motionplanning_1} utilizes attractive forces toward goals and repulsive forces from obstacles to generate smooth trajectories. \cite{motionplanning_2,liu2024} refined the original concept by introducing n-ellipse representations that vary with distance, improving convergence properties. For dual-arm systems, APF methods face unique challenges related to workspace sharing and coordinated manipulation in manufacturing contexts. \cite{motionplanning_3} employed an alternating master-slave paradigm with charged surface workspace modeling for collision avoidance. However, the sequential trajectory planning approach lacks true coordination and may cause deadlocks in constrained workspaces. \cite{motionplanning_4} developed D-APF for UAV navigation using exponential fields with vertical motion prioritization, though it suffers from oscillations in complex obstacle environments. \cite{motionplanning_5} proposed hybrid APF-RRT combinations for mobile dual-arm systems, while \cite{motionplanning_6} integrated force sensing with rotating repulsive fields for enhanced navigation capabilities. 

Further learning methods, \cite{dual-arm, hao2024coordinated} proposed a hierarchical leader-follower DRL framework using Soft Actor-Critic(SAC) for dual-arm closed-chain manipulation in cooperative transportation tasks. While their approach demonstrates task adaptability to random target poses, it requires extensive offline training for convergence and the fixed leader-follower structure limits applicability, performing well in symmetric manipulation scenarios but struggling with asymmetric dual-arm coordination where flexible role switching is essential. Hierarchical approaches attempt to balance optimality and computational efficiency. \cite{intro6} introduced a two-layer framework using dynamic programming for policy trees and Conflict-Based Search (CBS) for inter-agent resolution. While effective for structured environments, the sequential planning then coordination approach lacks the proactive competitive evaluation and continuous proximity optimization capabilities needed for maximizing efficiency in dynamic dual-arm manipulation scenarios. Further for asymmetric manipulation like assembly operation \cite{alles2022learning} used modular approach with two decentralized single-arm controllers,  coupled using a single centralized learned policy. \cite{wang2025learning} developed a CNN-based proximal policy optimization (PPO) framework for dual-arm push-grasp synergy in dense clutter, generating adaptive 6-DoF grasps and flexible dual-arm push trajectories. \cite{jiang2023mastering} employed PPO with Transformer-based temporal reasoning for dual-arm assembly tasks through three-stage sim-to-real transfer using generative adversarial imitation learning (GAIL). However, their approach requires extensive simulation training and executes at only 2 Hz in the real world due to heavy CNN+Transformer inference, limiting its applicability to real-time manipulation tasks compared to reactive control methods. Despite their capabilities, 
DRL-based dual-arm methods share fundamental limitations: (1) substantial computational costs requiring parallel simulation 
environments for convergence, (2) complex sim-to-real transfer necessitated by physics discrepancies between simulation and physical 
systems, and (3) task-specific reward engineering where constraint modifications demand complete policy retraining. These limitations motivate reactive approaches that provide analytical solutions with real-time responsiveness and straightforward parameter adaptation without training overhead.

Both task allocation and motion planning rely critically on real-time environmental perception to meet industrial automation requirements for safety, throughput, and reliability. The integration of real-time vision feedback is essential for autonomous manufacturing systems, requiring real-time perception for reactive behavior \cite{santos2024}. Visual perception must directly drive robot movement and task completion without human intervention~\cite{vision_1}. \cite{vision_2} demonstrated vision-based template matching approaches using CAD model correspondence for industrial components, achieving robust sorting performance through geometric feature extraction and size-based classification in cluttered workspaces. However, existing vision-integrated dual-arm systems typically employ offline planning approaches that cannot adapt to rapidly changing object configurations in dynamic manufacturing environments.

Current dual-arm systems face critical deployment limitations: (1) market-based methods optimize temporal task sequencing through auction-based allocation but lack spatial conflict resolution, resulting in collision risks during simultaneous operation in shared workspaces, (2) hierarchical approaches lack continuous proximity optimization for efficient workspace utilization, (3) computationally intensive methods requiring complete workspace pre-computation, (4) DRL-based methods require intensive parallel training, complex sim-to-real transfer, and complete policy retraining for constraint modifications, limiting real-time industrial applicability, and (5) absence of integrated mechanisms addressing both temporal sequencing and spatial collision avoidance for simultaneous multi-object handling scenarios.

This work addresses these limitations by introducing a novel intelligent bucket management and task allocation framework that dynamically optimizes object assignment based on real-time proximity analysis. Unlike conventional immediate assignment upon detection, our approach implements temporal proposal windows that enable competitive evaluation of multiple objects over five detection frames. The key innovation lies in the distance-competitive selection mechanism that continuously updates target assignments when significantly closer objects are detected, while maintaining system stability through proposal limits and spatial conflict resolution. The framework achieves  ($O(n\cdot|O|)$) linear complexity, enabling real-time execution without workspace pre-computation requirements.

\noindent \textbf{Key Technical Contributions:}
\begin{enumerate}
    \item \textbf{Temporal Proposal Windows:} Each arm evaluates up to five consecutive vision frames before committing to optimal distance-based selection, enabling superior task allocation decisions.
    \item \textbf{Distance-Competitive Updates:} Real-time target replacement when significantly closer objects are detected during evaluation phases, optimizing workspace utilization.
    \item \textbf{Spatial Conflict Resolution:} Prevents multiple arms from targeting identical objects through proximity validation with 0.01m threshold accuracy.
    \item \textbf{Integrated Vision-Motion Pipeline:} Seamless integration between YOLOv8 OBB object detection on Realsense D415 at 30FPS, dynamic task allocation, and real-time motion planning at 10Hz.
\end{enumerate}
\section{Methodology}
\label{method}
\begin{figure}[!t]
\centering
\includegraphics[width=\linewidth]{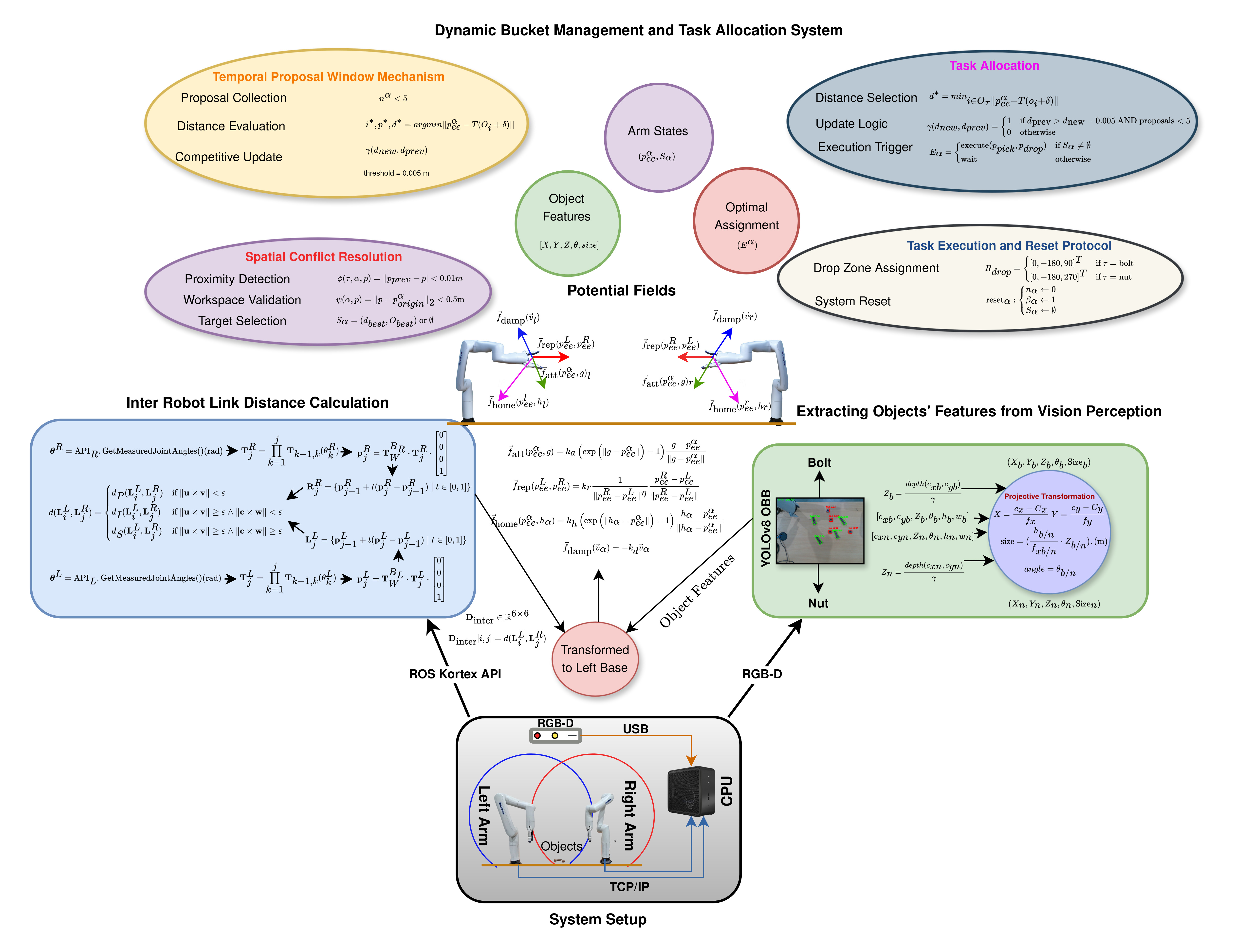}
\caption{Overall Methodology: Intelligent Task Allocation for Dual Arm Efficient Motion Planning to Handle Objects.}
\label{fig:overall_method}
\end{figure}

The overall methodology integrates four key components for efficient dual-arm motion planning with intelligent task allocation, as illustrated in Fig.~\ref{fig:overall_method}. First, inter-arm link distance calculation through continuous monitoring of joint configurations and end-effector positions using real-time forward kinematics from ROS Kortex API feedback. Second, vision-based object detection and pose estimation provides real-time workspace monitoring using YOLOv8 OBB for simultaneous object classification and orientation estimation. Third, dynamic bucket management system serves as the central coordinator, employing proximity-based competitive task allocation through temporal proposal windows and spatial conflict resolution. Fourth, an improved Artificial Potential Field (iAPF) framework implements coordinated dual-arm movements through goal attraction, inter-arm repulsion, home-seeking, and velocity damping force fields.
\subsection*{System Architecture and Experimental Setup}
\begin{figure}[h!]
    \centering
    \includegraphics[width=0.5\linewidth]{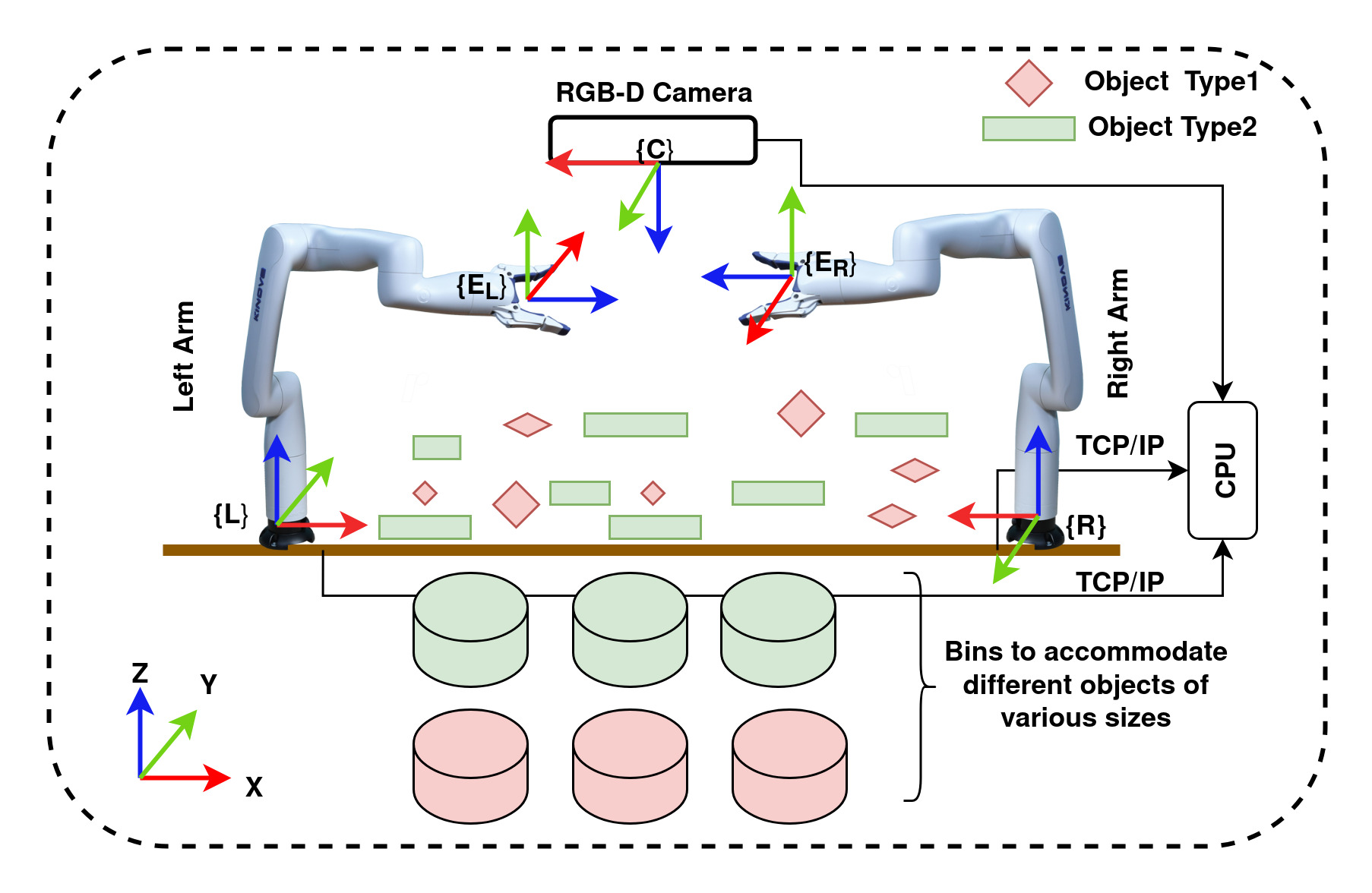}
    \caption{Dual-Arm setup for intelligent task allocation}
    \label{Experimentalsetup}
\end{figure}

The system configuration for dual-arm motion planning, as shown in Fig.~\ref{Experimentalsetup}, consists of two Kinova Gen3 Lite 6 DOF  manipulators and an Intel Realsense D415 RGB-D camera. The processing unit is an Intel NUC9 with 32GB RAM and an RTX 3060 for graphical processing. The dual-arm system mounted on a fixed base with a separation of 0.761 m. The globally fixed overhead RGB-D camera provides global workspace monitoring at 640$\times$480 resolution with depth sensing capability. The manipulators interface via TCP/IP sockets using the Kortex API, while camera connects through USB 3.0 for real-time data publishing at 30 FPS using ROS1(Noetic) communication.
\begin{equation}
T_{LR} = 
\begin{bmatrix}
-1 & 0 & 0 & 0.761 \\ 
0 & -1 & 0 & 0 \\ 
0 & 0 & 1 & 0 \\ 
0 & 0 & 0 & 1
\end{bmatrix}
\label{TLR}
\end{equation}

\begin{equation}
T_{RC} = 
\begin{bmatrix}
1 & 0 & 0 & 0.250 \\ 
0 & -1 & 0 & -0.210 \\ 
0 & 0 & -1 & 0.75 \\ 
0 & 0 & 0 & 1
\end{bmatrix}
\label{TRC}
\end{equation}

\begin{equation}
    T = T_{LR}\cdot T_{RC}
    \label{baseeq}
\end{equation}
\begin{equation}
T_{LE_{L}} = \prod_{i=1}^{6} T_{i-1,i}(\theta_i^L) \cdot T_{6,E_{L}}
\label{left_fk}
\end{equation}

\begin{equation}
T_{RE_{R}} = \prod_{i=1}^{6} T_{i-1,i}(\theta_i^R) \cdot T_{6,E_{R}}
\label{right_fk}
\end{equation}

\begin{equation}
T_{LE_{L}} = I_4 \quad \text{(Left arm in its own frame)}
\end{equation}

\begin{equation}
T_{LE_{R}} = T_{LR} \cdot T_{RE_{R}} \quad \text{(Right arm transformed to left frame)}
\end{equation}
\begin{align}
\label{left_frame}
T_{LE_{L}} &= \prod_{i=1}^{6} T_{i-1,i}(\theta_i^L) \cdot T_{6, E_{L}}  \\
T_{LE_{R}} &= T_{LR} \cdot \prod_{i=1}^{6} T_{i-1,i}(\theta_i^R) \cdot T_{6, E_{R}}
\label{right_frame}
\end{align}
$\theta_{i}^{L}, \theta_{i}^{R}$ are joint angles for left/right arms measured from Kortex API feedback.\\
$T_{i-1,i}(\theta_{i})$ are individual joint transformations.\\
$T_{6,E_{L}}, T_{6,E_{R}}$ are end-effector transformations.\\
$T_{LR}$ is the left-to-right base transformation.

\noindent Continuous current end-effector poses given in Eq.~\ref{left_frame} and Eq.~\ref{right_frame} updated at 10 Hz through real-time joint angle measurements from the Kortex API. The vision-control integration operates with 30 FPS detection and 10 Hz control loops, synchronized via ROS queue size =1 to access the most recent frame with maximum 70ms latency.

\noindent The coordinate frames {L}, {R}, {C}, {$E_{L}$}, and {$E_{R}$} represent left arm, right arm base frames, camera, and end effector frames of left and right frames respectively. A global left-base {L} frame is considered for unified control, the transformation $T_{LR}$ and $T_{RC}$ as given by Eq.~\ref{TLR} and Eq.~\ref{TRC} enable seamless coordinate conversions between different reference frames, ensuring consistent spatial relationships throughout the system. The control loop operates at 10 Hz frequency, providing real-time responsiveness for dynamic task allocation and collision avoidance while maintaining system stability during complex dual-arm manipulation tasks.

\subsection{Real-Time Inter-Arm Link Distance Monitoring}
For inter-arm collision avoidance in dual-arm manipulator systems, continuous real-time monitoring of inter-arm distances is essential. The forward kinematics established in Eq.~\ref{left_frame} and Eq.~\ref{right_frame} enables modeling each manipulator link as a line segment in 3D space for precise distance calculations. The global position of joint $j$ on arm $\alpha$ is:
\begin{equation}
    p^{\alpha}_{j} = T^{\alpha}_{E\alpha} \cdot [0, 0, 0, 1]^{T} \in \mathbf{R^{3}}
\end{equation}
where $\alpha \in \{L, R\}$ represents left/right arm.

\noindent Each manipulator link is represented as a line segment:
\begin{equation}
    S^{\alpha}_{j} = \{p^{\alpha}_{j-1} + t(p^{\alpha}_{j}-p^{\alpha}_{j-1}) | t\in[0,1]\}
\end{equation}

\noindent The complete segment set for dual-arm system:
\begin{equation}
    \mathcal{S} = \{S^{L}_{j}, S^{R}_{j} | j\in[1,6]\}
\end{equation}
\noindent The inter-arm distance matrix $D_{inter} \in R^{6\times 6}$ is defined as:
\begin{equation}
    D_{inter}[i, j] = d(S^{L}_{i}, S^{R}_{j}) \quad \text{where} \quad i, j \in [1, 6]
\end{equation}

\noindent The critical collision metric is:
\begin{equation}
    d_{min} = \text{min}(D_{inter})
    \label{min_dist}
\end{equation}

This real-time distance monitoring provides the foundation for collision avoidance in the subsequent iAPF framework, minimum distance Eq.~\ref{min_dist} directly influences repulsive forces magnitudes between manipulators.
\subsection{Vision Based Object Detection and Pose Estimation}
Real-time vision perception with high precision is essential for effective task allocation, manipulation, and accurate size estimation of industrial components. YOLOv8 OBB \cite{yolov8} provides both object classification and orientation estimation through its oriented bounding box. This eliminates computational overhead of separate detections and pose estimations \cite{shiporientation}. Custom dataset prepared with 1230 images with diverse real-world conditions including varying backgrounds, lighting, and object ages(new and old components). Further dataset increased to tenfold using image augmentation techniques. The model trained for 250 epochs on prepared custom dataset. Training employed standard hyper-parameters with early stopping based on validation loss convergence. The confusion matrix as shown in Table~\ref{tab:confusion_matrix} demonstrates excellent discrimination between bolt (0.99) and nut (0.97) classes with minimal inter-class confusion (0.02). Further the vision implementation parameter details are listed in Table~\ref{tab:visionparameters}.

\begin{table}[t]
\centering
\caption{Normalized Confusion Matrix for YOLOv8 OBB Model (250 epochs)}
\begin{tabular}{ccccc}
\toprule
\textbf{Predicted} & \textbf{Bolt} & \textbf{Nut} & \textbf{Reference} & \textbf{Background} \\
\midrule \midrule
\textbf{Bolt} & 0.99 & 0.02 & 0.00 & 0.75 \\
\textbf{Nut} & 0.00 & 0.97 & 0.00 & 0.25 \\
\textbf{Reference} & 0.01 & 0.02 & 0.42 & 0.00 \\
\textbf{Background} & 0.00 & 0.00 & 0.57 & 0.00 \\
\bottomrule
\end{tabular}
\label{tab:confusion_matrix}
\end{table}

\noindent Following object classification, OBB centre coordinates ($c_{x}, c_{y}$) are transformed from image coordinates to 3D Cartesian coordinates using camera intrinsic parameters ($C_{x}, C_{y}, f_{x}, f_{y}$) and depth information ($Z$). The projective transformation \cite{projective_transformation} employs the standard pinhole camera model:
\begin{equation}
\begin{aligned}
Z &= \frac{\text{depth}(c_x, c_y)}{\gamma} \\[0.3em]
X &= \frac{(c_x - C_x) \cdot Z}{f_x} \\[0.3em]
Y &= \frac{(c_y - C_y) \cdot Z}{f_y} \\[0.3em]
size &= \frac{(h \cdot Z) \cdot \gamma}{f_x}, \quad angle &= \theta
\end{aligned}
\label{vision_transform}
\end{equation}
where $\gamma$ is scaling factor.
The detection noise is filtered through model's confidence score rejecting uncertain detections. Depth validation maintains last valid measurement when sensor reads zero. Two-stage angle correction resolves OBB ambiguity (aspect ratio check) and maps to achievable gripper orientations. Further the implementing details are listed in Table~\ref{tab:visionparameters}.

\noindent Further, object information transformed to left base arm as shown in Eq.~\ref{transform},
\begin{equation}
    T_{LR} \cdot T_{RC} \cdot [X, Y, Z, 1]^{T}
    \label{transform}
\end{equation}

\noindent The transformed object coordinates enable proximity-based competitive selection within the temporal proposal framework described next.
\begin{table}[t]
    \centering
    \caption{{Vision-control integration and task allocation parameters}}
    \label{tab:visionparameters}
    \scriptsize
    \begin{tabular}{lcc}
        \toprule
        \textbf{Parameter} & \textbf{Value} & \textbf{Description} \\
        \midrule\midrule
        YOLOv8 inference rate & $>$70 FPS  & RTX 3060 GPU \\
        Camera acquisition rate & $\approx$ 30 FPS & RSD415 hardware limit at (640$\times$480) \\
        Control loop rate & 10 Hz    & Dual-arm coordination frequency \\
        Detection confidence & 0.75 & YOLOv8 threshold (configurable) \\
        ROS queue size & 1 & Latest-frame synchronization \\
        Batch Size & 1 & Real-time single-frame processing\\
        Proposal window & 5 frames & Temporal evaluation period (Eq.\ref{eq:update_logic}) \\
        Distance improvement & 0.005 m & Competitive update threshold (Eq.\ref{eq:update_logic}) \\
        Spatial conflict & 0.01 m & Target duplication prevention (Eq.\ref{eq:conflict_detection})  \\
        Reachability radius & 0.5 m     & Workspace constraint (Eq.\ref{eq:workspace_validation}) \\
        \midrule
    \end{tabular}
    \vspace{1mm}
    \begin{minipage}{0.98\columnwidth}
        \footnotesize \emph{Note:} The 30 FPS inference rate with 10 Hz control and queue size = 1 results in maximum 70ms vision-to-action latency.
    \end{minipage}
\end{table}
\subsection{Distance-based Competitive Task Allocation}
Traditional dual-arm systems utilize fixed task allocations where individual arms handle predetermined object types, with robot arms assigned to specific processing tasks and limited adaptability to dynamic object distributions. The dynamic task allocation framework optimizes object assignment through proximity-based competitive selection within temporal proposal windows. The system evaluates multiple object proposals over five detection frames, enabling optimal allocation decisions based on distance analysis and workspace constraints Table~\ref{temporal_window}.

\subsubsection*{\textbf{Proximity Based Object Selection}}
For each detected object set of type $\tau \in \{\text{nut}, \text{bolt}\}$ and arm $\alpha \in \{\text{left}, \text{right}\}$, the system performs proximity-based selection using:
\begin{equation}
i^*, p^*, d^* = \arg \min_{i \in O_\tau} \|p_{ee}^\alpha - T([x_i, y_i, z_i, 1]^T)_{1:3}\|_2
\label{eq:distance_selection}
\end{equation}
where $T$ represents the camera-to-base coordinate transformation Eq.~\ref{baseeq}, $[x_i, y_i, z_i]^T$ denotes the 3D position of object $i$, $O_{\tau}$ is the set of detected objects of type $\tau$, and $p_{ee}^\alpha$ represents the current end-effector position of arm $\alpha$.
\noindent Objects must satisfy the reachability constraint to ensure feasible task assignments:
\begin{equation}
\psi(\alpha, p) =
\begin{cases}
1 & \text{if } \|p - p_{origin}^\alpha\|_2 < 0.5\text{m} \\
0 & \text{otherwise}
\end{cases}
\label{eq:workspace_validation}
\end{equation}
where $p_{origin}^\alpha$ denotes the base origin position of arm $\alpha$, ensuring that proposed objects lie within the manipulator's operational workspace of $0.5~m$ radius.

\subsubsection*{\textbf{Conflict Resolution and Target Management}}
To prevent both arms from targeting the same object, the system implements proximity-based conflict detection that identifies when arms attempt to assign identical or nearby targets. The spatial conflict detection mechanism is defined as:
\begin{equation}
\phi(\tau, \alpha, p) =
\begin{cases}
1 & \text{if } \exists p_{prev}: \tau_{prev} = \tau \\
  & \text{AND } \|p_{prev} - p\|_2 < 0.01\text{m} \\
0 & \text{otherwise}
\end{cases}
\label{eq:conflict_detection}
\end{equation}
where $p_{prev}$ represents a previously assigned target position and $\tau_{prev}$ denotes its object type, with a spatial threshold of $0.01m$ ensuring precise conflict identification.

\noindent Each arm maintains a target store containing the best proposal from the current evaluation window:
\begin{equation}
S_\alpha = (d^{*}, O^{*}) \text{ or } \emptyset
\label{eq:target_store}
\end{equation}
where $d^{*}$ represents the optimal distance and $O^{*}$ contains the corresponding object information, or the store remains empty if no valid proposals exist.

\noindent When a new proposal arrives, the system evaluates whether to update the current target through competitive logic:
\begin{equation}
\gamma(d_{new}, d_{prev}) =
\begin{cases}
1 & \text{if } d_{\text{prev}} > d_{\text{new}} - 0.005\\ 
&\text{ AND proposals} < 5 \\
0 & \text{otherwise}
\end{cases}
\label{eq:update_logic}
\end{equation}
This mechanism(Eq.~\ref{eq:update_logic}) replaces the current target when the previous assignment is not significantly closer than the new proposal (within 0.005~m margin), preventing premature commitment while maintaining system stability through the proposal window limit.
\begin{table}[t]
    \centering
    \caption{Temporal Window Ablation Results}
    \label{temporal_window}
    \scriptsize
        \begin{tabular}{c c c c}
        \toprule
        \textbf{Window Size} & \textbf{Total Distance (m)} & \textbf{Total Time (s)} & \textbf{Avg.Velocity (m/s)}  \\
        \midrule\midrule
        1   & $16.44 \pm 0.32$ & $145.24 \pm 3.01$ & $0.113 \pm 0.0012$\\
        3   & $15.61 \pm 0.16$  & $132.87 \pm 2.801$ & $0.117 \pm 0.0013$\\
        \textbf{5*} & $\textbf{15.64} \pm \textbf{0.05}$ & $\textbf{130.51} \pm \textbf{1.54}$ & $\textbf{0.119} \pm \textbf{0.0012}$\\
        7   & $15.57 \pm 0.13$ & $144.72 \pm 2.49$& $0.107 \pm 0.0012$\\
        10  & $15.62 \pm 0.07$ & $160.22 \pm 0.48$&$0.097 \pm 0.0002$\\
        \bottomrule
        \end{tabular}%
    \vspace{2mm} 
    \begin{flushleft}
    \footnotesize *Selected configuration. Values: mean $\pm$ SD (n=5, 10 objects per trial)
    \end{flushleft}
\end{table}

\subsubsection*{\textbf{Task Execution and Reset Protocol}}
When the proposal window is exhausted, the system initiates task execution based on the availability of valid targets in each arm's store. The execution phase is governed by:
\begin{equation}
E_\alpha =
\begin{cases}
\text{execute}(p_{pick}, p_{drop}) & \text{if } S_\alpha \neq \emptyset \\
\text{wait} & \text{otherwise}
\end{cases}
\label{eq:execution_phase}
\end{equation}
where $p_{pick}$ represents the selected object position and $p_{drop}$ denotes the corresponding drop zone location.

\noindent Objects are assigned to predetermined drop zones based on object type to facilitate component sorting: 
\begin{equation} 
 R_{drop} = 
 \begin{cases} 
 [0, -180, 90]^T & \text{if } \tau = \text{bolt} \\ 
 [0, -180, 270]^T & \text{if } \tau = \text{nut} 
 \end{cases} 
\label{eq:drop_zone} 
\end{equation}

\noindent Upon task completion, all parameters reset to enable new proposal cycles and maintain system readiness for subsequent object detection:
\begin{equation} 
\text{reset}_\alpha : 
\begin{cases} n_\alpha \leftarrow 0 \\ 
\beta_\alpha \leftarrow 1 \\ 
S_\alpha \leftarrow \emptyset 
\end{cases} 
\label{eq:reset_protocol} 
\end{equation}
where $n_{\alpha}$ represents the proposal counter, $\beta_{\alpha}$ indicates the acceptance state, and $S_{\alpha}$ denotes the target store for arm $\alpha$. This reset mechanism ensures clean state transitions between allocation cycles while preserving system stability. The proposed framework achieves $\mathcal{O}(n_{\text{max}} \cdot |O|)$ computational complexity, where ${n_\text{max}} = 5$ represents the proposal window size and $|O|$ denotes detected objects, enabling real-time execution with 30 FPS vision processing and 10 Hz control loops.

\begin{figure}[!t]
\centering
\includegraphics[width=0.7\linewidth]{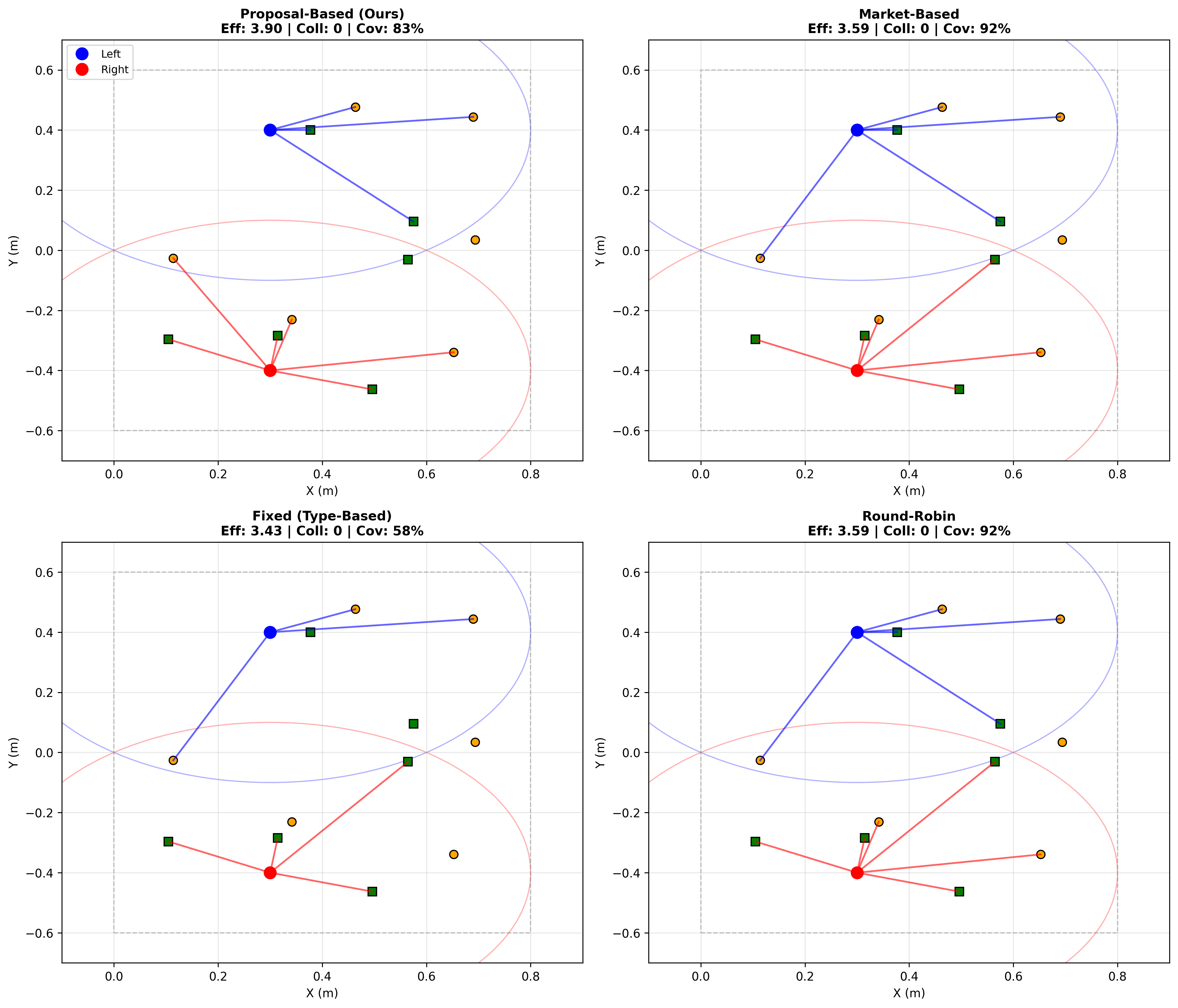}
\caption{Dual-arm workspace visualization comparing existing allocation methods with proposal method(our).}
\label{sim_setup}
\end{figure}

\begin{figure}[!t]
\centering
\includegraphics[width=\linewidth]{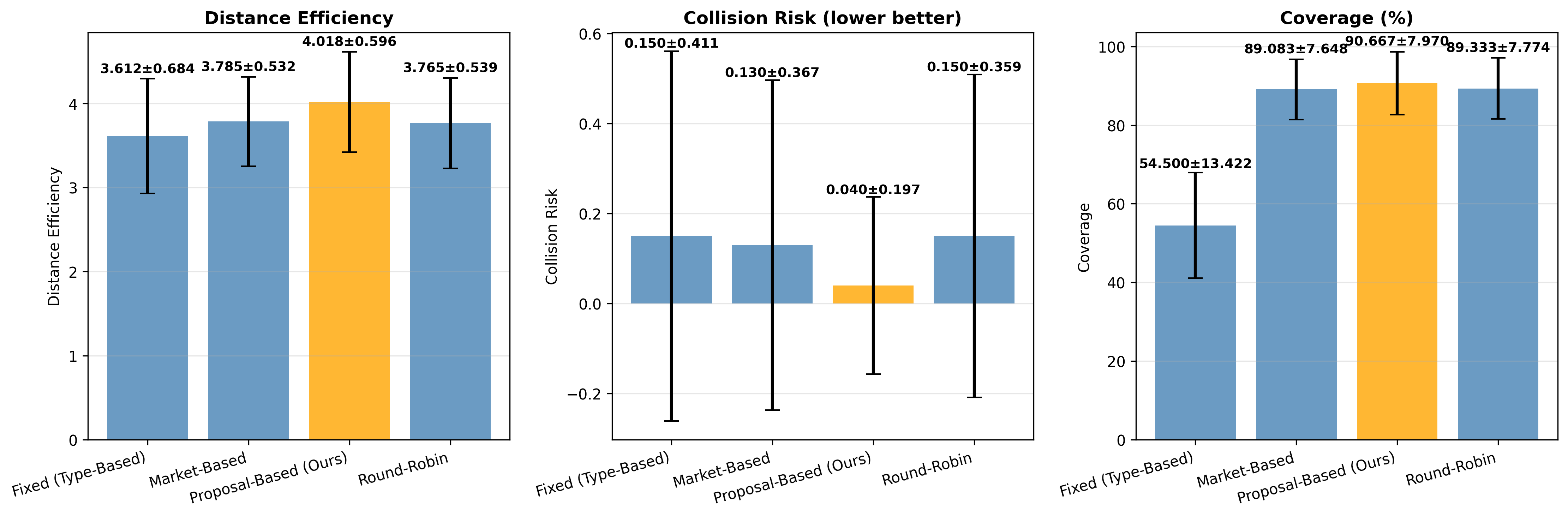}
\caption{Three performance metrics quantify allocation quality. [(1) Distance Efficiency 
(1/m) computed as the inverse of mean object-to-arm distance, measuring workspace utilization where higher values indicate better proximity-based assignments; 
(2) Collision Risk, representing the number of detected trajectory intersections during simultaneous arm operation; and (3) Coverage (\%), calculated as the percentage of successfully allocated objects relative to total objects.]}
\label{sim_metrics}
\end{figure}
\noindent The simulation environment implements dual-arm benchmarking, which closely resembles the experimental setup workspace as shown in Fig.~\ref{sim_setup}. This design ensures an unbiased evaluation of the four allocation strategies over 100 trials. Results in Fig.~\ref{sim_metrics} demonstrate the superiority of the proposed method across key metrics.

Proposal-based allocation achieves $4.020\pm0.595$ distance efficiency, outperforming market-based ($3.789\pm0.530$, 6.1\% improvement), round-robin ($3.769\pm0.537$, 6.7\% improvement), and fixed allocation ($3.618\pm0.681$, 11.1\% improvement). This improvement stems from competitive target updates within the five-frame proposal window: when a significantly closer object ($>0.005$m improvement) is detected, the arm replaces its previous target before commitment. Market-based methods lock decisions immediately following each auction, while round-robin ignores proximity entirely.

Most significantly, collision risk reduces by 69\% compared to market-based ($0.040\pm0.197$ vs $0.130\pm0.367$) and 73\% versus round-robin ($0.150\pm0.359$). Market-based sequential processing awards objects independently without checking if simultaneous assignments create trajectory intersections. Round-robin's alternating assignment without spatial awareness produces similar collision patterns.

Coverage remains consistent across competitive methods ($>89\%$) while fixed allocation achieves only $54.9\pm13.7\%$ due to rigid type constraints—right arm handles bolts, left handles nuts, leaving unreachable objects unassigned.

\begin{algorithm}
\caption{Temporal Proposal Window Mechanism}\label{algo2}
\begin{algorithmic}[1]
\State \textbf{Init:} $n_\alpha \leftarrow 0$, $\beta_\alpha \leftarrow \text{true}$, $S_\alpha \leftarrow \emptyset$
\For{each $(i, \tau, p, size)$ with arm $\alpha$ idle}
    \State $(d^*, p^*) \leftarrow \arg\min_{i \in \mathcal{O}_\tau} \|p^{\alpha}_{ee} - p_i\|_2$
    \If{$\psi(\alpha, p^*) = 1 \wedge \phi(\tau, \alpha, p^*) = 0$}
        \If{$n_\alpha < 5$}
            \If{$S_\alpha = \emptyset \vee d_{\text{prev}} > d^* - 0.005$}
                \State $S_\alpha \leftarrow (d^*, \text{Object}(\tau, size, p^*))$
            \EndIf
            \State $n_\alpha \leftarrow n_\alpha + 1$
        \Else
            \State $\beta_\alpha \leftarrow \text{false}$
        \EndIf
    \EndIf
\EndFor
\If{$\beta_\alpha = \text{false} \wedge S_\alpha \neq \emptyset$}
    \State \textbf{Execute:} $(p_{\text{pick}}, p_{\text{drop}}) \leftarrow (S_\alpha.\text{pick}, S_\alpha.\text{drop})$
    \State \textbf{Reset:} $n_\alpha \leftarrow 0$, $\beta_\alpha \leftarrow \text{true}$, $S_\alpha \leftarrow \emptyset$
\EndIf
\end{algorithmic}
\end{algorithm}
\subsection{Improved Artificial Potential Field Framework}
The proposed iAPF framework from our previous work \cite{iAPF}, employs a three-force equilibrium system for coordinated dual-arm manipulation. Let $p_{ee}^{\alpha}$ denote the end-effector position of arm $\alpha \in \{\text{left}, \text{right}\}$ in the unified left-base frame, and $g$ represent the target-object position transformed from camera coordinates to the left-base frame using transformation $T$ Eq.~\ref{baseeq}. The improved APF framework uses the specific force constants empirically tuned for the Kinova Gen3 Lite Manipulators as shown in Table~\ref{tab:parameters}.

\subsubsection{Force Field Definitions}
\begin{figure}
    \centering
    \includegraphics[width=0.7\linewidth]{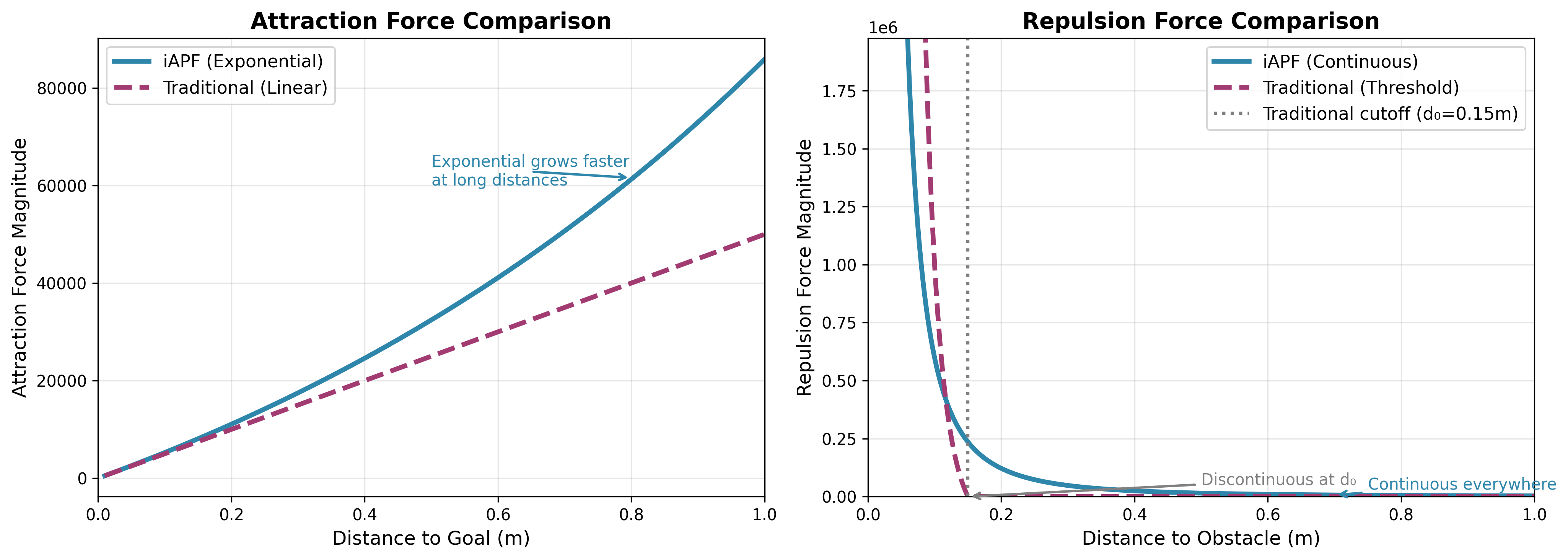}
    \caption{Force profile comparison showing iAPF advantages: Exponential attraction with adaptive gradient(left), Continuous repulsion eliminating traditional threshold discontinuity(right).}
    \label{comp_forces}
\end{figure}

The improved APF framework defines four coordinated force components for dual-arm manipulation. The goal attraction force provides exponential convergence characteristics:
\begin{equation}
    \vec{f}_{\text{att}}(p_{ee}^{\alpha}, g) = 
k_a \left(e^{\lVert g - p_{ee}^{\alpha}\rVert} - 1\right) 
\frac{g - p_{ee}^{\alpha}}{\lVert g - p_{ee}^{\alpha}\rVert}    
\end{equation}

\noindent the exponential attraction forces provide rapid convergence near targets as shown in Fig.\ref{comp_forces}(left), the exponential attraction (blue continuous) provides sublinear growth near goal ($d<0.2m$) for gentle convergence and superlinear growth at far distances ($d>0.5m$) for faster approach, eliminating the need for gain scheduling. Traditional linear attraction(red dotted) lacks this adaptive behavior.

\noindent The inter-arm repulsive force ensures continuous collision avoidance through inverse-distance relationships:
\begin{equation}
\vec{f}_{\text{rep}}(p_{ee}^L, p_{ee}^R) = k_r \frac{1}{d_{min}^{\eta}} \frac{p_{ee}^R - p_{ee}^L}{\lVert p_{ee}^R - p_{ee}^L\rVert}
\label{eq:repulsive_force}
\end{equation}

this formulation provides continuous repulsion at all distances, enabling smoother dual-arm coordination as shown in Fig.~\ref{comp_forces}(right). Continuous repulsion (blue continuous) maintains smooth force decay across all distances, while traditional threshold-based repulsion(red dotted) creates discontinuity at $d_{0}$=0.15 m, causing jerks and oscillations that are not suitable for dual-arm coordination.

\noindent The home-seeking force guides arms toward predefined safe positions when not actively engaged:
\begin{equation}
\vec{f}_{\text{home}}(p_{ee}^{\alpha}, h_{\alpha}) = 
k_h \left(e^{\lVert h_{\alpha} - p_{ee}^{\alpha}\rVert} - 1\right) 
\frac{h_{\alpha} - p_{ee}^{\alpha}}{\lVert h_{\alpha} - p_{ee}^{\alpha}\rVert}
\label{eq:home_force}
\end{equation}
Velocity damping ensures system stability by opposing motion:
\begin{equation}
\vec{f}_{\text{damp}}(\vec{v}_{\alpha}) = -k_d \vec{v}_{\alpha}
\label{eq:damping_force}
\end{equation}
where $k_a$, $k_r$, $k_h$, and $k_d$ are positive gain constants, and $\eta$ represents the decay exponent for repulsive forces listed in Table~\ref{tab:parameters}.
The combined force is shown in Eq.~\ref{combined_forces}
\begin{equation}
    \vec{f}_{combined} = c_{\text{goal}}\,\vec{f}_{\text{att}} + c_{\text{obs}}\,\vec{f}_{\text{rep}} + \vec{f}_{\text{home}} + \vec{f}_{\text{damp}}
    \label{combined_forces}
\end{equation}
The selection of the APF parameters, involved systematic ablation studies by independently varying each coefficient. Table~\ref{cobs_stats} and Table~\ref{cgoal_stats} present the comparative analysis over 40 trials. Each trial involved 10 objects (5 nuts and 5 bolts) randomly and compactly distributed within the shared workspace to replicate realistic industrial sorting scenarios. As shown in Table~\ref{cobs_stats}, $C_{\text{goal}}$=0.05 achieved stable but inefficient operation with a completion time of $\approx$170 s, while $C_{\text{goal}}$=0.15 exhibited near-instability and oscillations during simultaneous pick-and-place operations. $C_{\text{goal}}$=0.2 induced aggressive motion, resulting in ground(table) collisions and system(arm) faults.

\noindent The repulsion coefficients in Table~\ref{cgoal_stats}, $C_{\text{obs}}$=0.005 demonstrated conservative behavior with significantly closer inter-arm proximity. To achieve a balance between safety and efficiency, $C_{\text{obs}}$=0.01 was selected. Higher values, such as $C_{\text{obs}}$=0.015, induced oscillatory motion and extended completion times. Consequently, the optimal parameters were determined to be $C_{\text{goal}}$=0.1 and $C_{\text{obs}}$=0.01, achieving a completion time of $\approx$130 s with zero collisions across all trials.
\subsubsection{Resultant Force and Velocity Generation}
The resultant force combines all components with task-adaptive coefficients:
\begin{equation}
\vec{f}_{\text{resultant}} = \frac{1}{s}\left(\vec{f}_{combined}\right)
\label{eq:resultant_force}
\end{equation}
$s$ is a normalization scaling factor, to prevent actuator saturation by bounding the combined force magnitude as shown in Table~\ref{tab:parameters}.

\noindent Linear velocities are generated through force-to-velocity conversion:
\begin{equation}
\vec{v}_{\alpha} = \frac{\vec{f}_{\text{resultant}}^{\alpha}}{s_v}
\label{eq:velocity_generation}
\end{equation}
where $s_v$ represents the force-to-velocity scaling factor from Table~\ref{tab:parameters} for maintaining stable system dynamics while ensuring responsive trajectory execution compatible with the Kortex API . The generated velocities are transformed to respective base frames:
\begin{equation}
\vec{v}_{L} = \frac{\vec{f}_{\text{resultant}}^{L}}{s_v}, \quad \vec{v}_{R}^{\text{right}} = R_{LR} \cdot \frac{\vec{f}_{\text{resultant}}^{R}}{s_v}
\label{eq:velocity_transform}
\end{equation}
\begin{figure}
    \centering
    \includegraphics[width=0.7\linewidth]{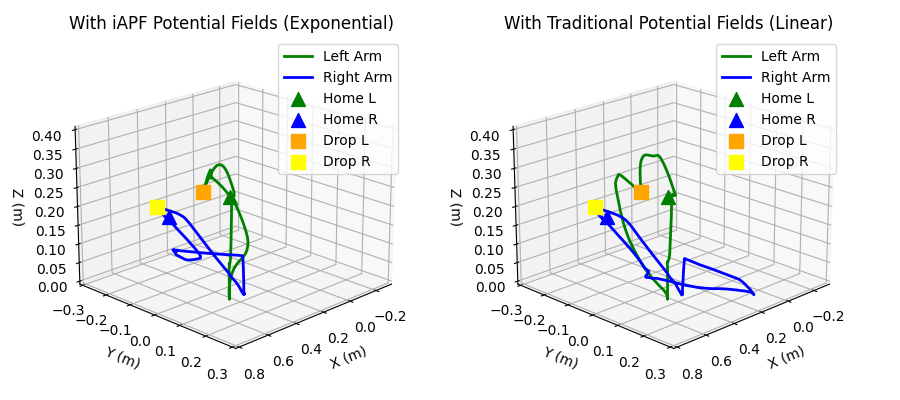}
    \caption{Dual-Arm Trajectories comparison with iAPF force(left) and with traditional APF(right).}
    \label{trajec_comp}
\end{figure}

\begin{figure}
    \centering
    \includegraphics[width=0.7\linewidth]{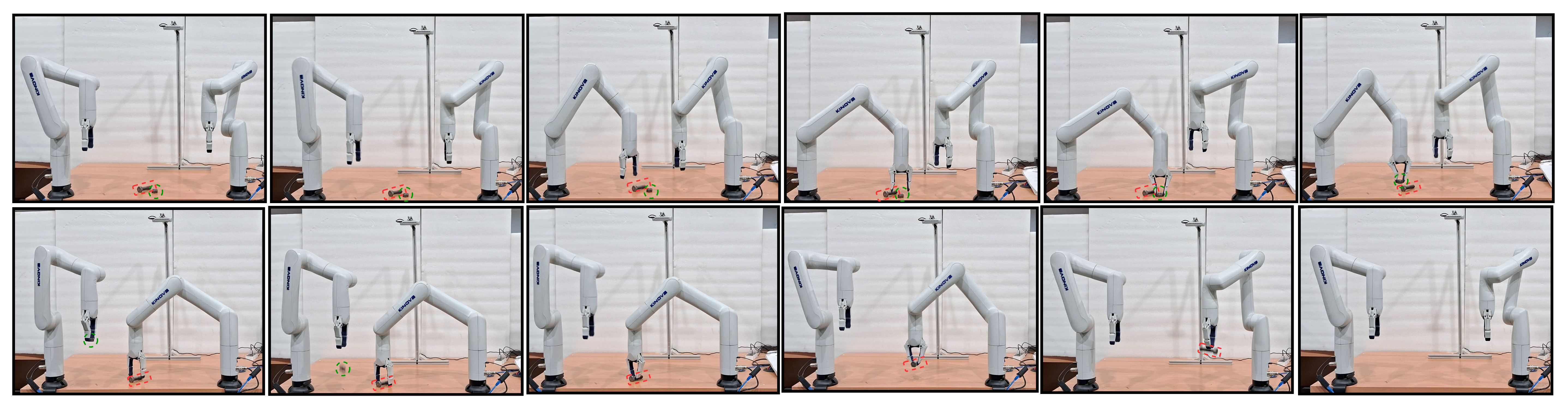}
    \caption{Dual-Arm sequential movements with iAPF handling conflicting near-goal industrial components.}
    \label{sequential}
\end{figure}

\noindent The system dynamically adjusts the force coefficients based on the manipulation phases. During targeting, collision avoidance operates with $c_{\text{obs}} > 0$, while priority locking assigns precedence to the arm with a shorter target distance. For precision gripping, collision avoidance is disabled ($c_{\text{obs}} = 0$) to prevent interference with accurate positioning.

\noindent Orientation control integrates object orientations from vision with real-time forward kinematics. Current(Eq.~\ref{left_fk}, \ref{right_fk}) and desired orientations $\in$ SO(3) are defined as:
\begin{equation}
\begin{aligned}
R_c^\alpha = (R_{LE_{L}}) \text{ or } (R_{LE_{R}}) \\
R_{d}^{\alpha} = R_z(\theta_z) \cdot R_y(\theta_y) \cdot R_x(\theta_x)
\label{eq:orientations}
\end{aligned}
\end{equation}
where $\theta_{d}^{\alpha} \in \mathbf{R}^{3}$ represents the desired end-effector orientation for arm $\alpha$ in the left base frame, with $\theta_{x} = 0^{\circ}$, $\theta_{y} = -180^{\circ}$ (downward-pointing gripper), and $\theta_{z}$ obtained from Eq.~\ref{vision_transform} after transforming the detected object angle to the robot base coordinate system.

\noindent The orientation error and the angular velocity control follow:
\begin{equation}
R_e^\alpha = R_d^\alpha \cdot (R_c^\alpha)^{-1}
\label{eq:orientation_error}
\end{equation}
the orientation error $R_{e}^{\alpha}$ is expressed in axis-angle representation as $\mathbf{\Theta} = \Theta \hat{\mathbf{k}}$, where $\hat{\mathbf{k}} \in \mathbf{R^{3}}$ is the unit rotation axis and $\Theta$ is the rotation magnitude.

\begin{equation}
\vec{\omega}_\alpha = K_p \cdot \mathbf{\Theta} + K_d \cdot \frac{d}{dt}(\mathbf{\Theta})
\label{eq:angular_control}
\end{equation}

Angular velocities are transformed into respective arm frames:
\begin{equation}
\vec{\omega}_L = \vec{\omega}_L^{\text{left}}, \quad \vec{\omega}_R = R_{LR} \cdot \vec{\omega}_R^{\text{left}}
\label{eq:angular_transform}
\end{equation}
where $K_p, K_d \in \mathbf{R}^{3 \times 3}$ are diagonal gain matrices with elements $k_p = 30$ and $k_d = 0.1$, both linear and angular velocities (Eq.~\ref{eq:velocity_transform}, \ref{eq:angular_transform}) for proper dual-arm control.
        
        
        
\begin{table}[h!]
    \centering
    \caption{$C_{\text{goal}}$ Parameter Validation ($C_{\text{obs}}$ = 0.01) [Y: Yes N: No]}
    \label{cobs_stats}
    \scriptsize
    \begin{tabular}{c c c c c c}
    \toprule
    \textbf{$C_{\text{goal}}$} & 
    \textbf{\parbox{1.8cm}{\centering Completion\\Dist. (m)}} & 
    \textbf{\parbox{1.8cm}{\centering Completion\\Time (s)}} & 
    \textbf{\parbox{1.8cm}{\centering Avg.Velocity\\(m/s)}} & 
    \textbf{Collisions} & 
    \textbf{Overshoot} \\
    \midrule\midrule
    0.05 & $15.50 \pm 0.109$ & $170.88 \pm 0.32$ & $0.090 \pm 0.001$ & N & N \\
    \textbf{0.10*} & $\textbf{15.65} \pm \textbf{0.057}$ & $\textbf{130.51} \pm \textbf{1.54}$ & $\textbf{0.119} \pm \textbf{0.0012}$ & \textbf{N} & \textbf{N} \\
    0.15 & $16.67 \pm 0.138$ & $119.64 \pm 0.86$ & $0.139 \pm 0.0003$ & N & Y \\
    0.2  & - & - & - & Y & - \\
    \bottomrule
    \end{tabular}
\end{table}
        
        
        
        
\begin{table}[h!]
    \centering
    \caption{$C_{\text{obs}}$ Parameter Validation ($C_{\text{goal}}$ = 0.1) [Y: Yes N: No]}
    \label{cgoal_stats}
    \scriptsize
    \begin{tabular}{c c c c c c}
    \toprule
    \textbf{$C_{\text{obs}}$} & 
    \textbf{\parbox{1.8cm}{\centering Completion\\Dist. (m)}} & 
    \textbf{\parbox{1.8cm}{\centering Completion\\Time (s)}} & 
    \textbf{\parbox{1.8cm}{\centering Avg.Velocity\\(m/s)}} & 
    \textbf{Collisions} & 
    \textbf{Overshoot} \\
    \midrule\midrule
    0.005 & $13.87 \pm 0.103$ & $103.92 \pm 3.89$ & $0.133 \pm 0.003$ & N & N \\
    \textbf{0.010*} & $\textbf{15.65} \pm \textbf{0.057}$ & $\textbf{130.51} \pm \textbf{1.54}$ & $\textbf{0.119} \pm \textbf{0.0012}$ & \textbf{N} & \textbf{N} \\
    0.015 & $17.26 \pm 0.172$ & $143.38 \pm 0.69$ & $0.120 \pm 0.0013$ & N & Y \\
    0.02  & $18.46 \pm 0.165$ & $145.25 \pm 0.57$ & $0.127 \pm 0.001$ & N & Y \\
    \bottomrule
    \end{tabular}
\end{table}

\noindent Fig.\ref{trajec_comp} compares the dual-arm trajectories in scenarios involving near-goal conflicts. The iAPF framework (left) demonstrates smooth and stable convergence for both robot arms as shown in Fig.~\ref{sequential}, whereas the traditional APF (right) exhibits instability. Specifically, the trajectory of the right arm (blue) deviates sharply when the repulsive force spikes due to the close proximity and simultaneous operation of the two arms. The enhanced stability observed in the iAPF framework is attributed to two key mechanisms: the home-seeking attraction force effectively balances the repulsion force spikes, and the task-adaptive coefficients successfully eliminate inter-arm collisions.
\begin{table}[t]
    \centering
    \caption{Implementation Parameters for the Improved Artificial Potential Field Framework}
    \label{tab:parameters}
    \scriptsize
    \begin{tabular}{lcc}
        \toprule
        \textbf{Parameter} & \textbf{Value} & \textbf{Description} \\
        \midrule\midrule
        Attraction constant ($k_{a}$)    & 50{,}000   & Exponential goal attraction gain \\
        Repulsion constant ($k_{r}$)     & 3{,}000    & Inter-arm repulsion gain \\
        Home-seeking constant ($k_{h}$)  & 600{,}000  & Home position attraction gain \\
        Damping coefficient ($k_{d}$)    & 100        & Velocity damping gain \\
        Repulsion exponent ($\eta$)      & 2      & Inverse-distance decay rate \\
        Normalization factor ($s$)       & 4          & Combined force scaling \\
        Velocity scaling ($s_{v}$)       & 1{,}000    & Force-to-velocity conversion \\
        Goal coefficient ($c_{\text{goal}}$) & 0.1     & Attraction weight \\
        Obstacle coefficient ($c_{\text{obs}}$) & 0.01 & Repulsion weight \\
        \midrule
    \end{tabular}
    \vspace{1mm}
    \begin{minipage}{0.98\columnwidth}
        \footnotesize \emph{Note:} Force constants and scaling factors were empirically tuned for Kinova Gen3 Lite manipulators to balance convergence speed and collision safety.
    \end{minipage}
\end{table}

\begin{figure}[h!]
    \centering
    \includegraphics[width=0.7\linewidth]{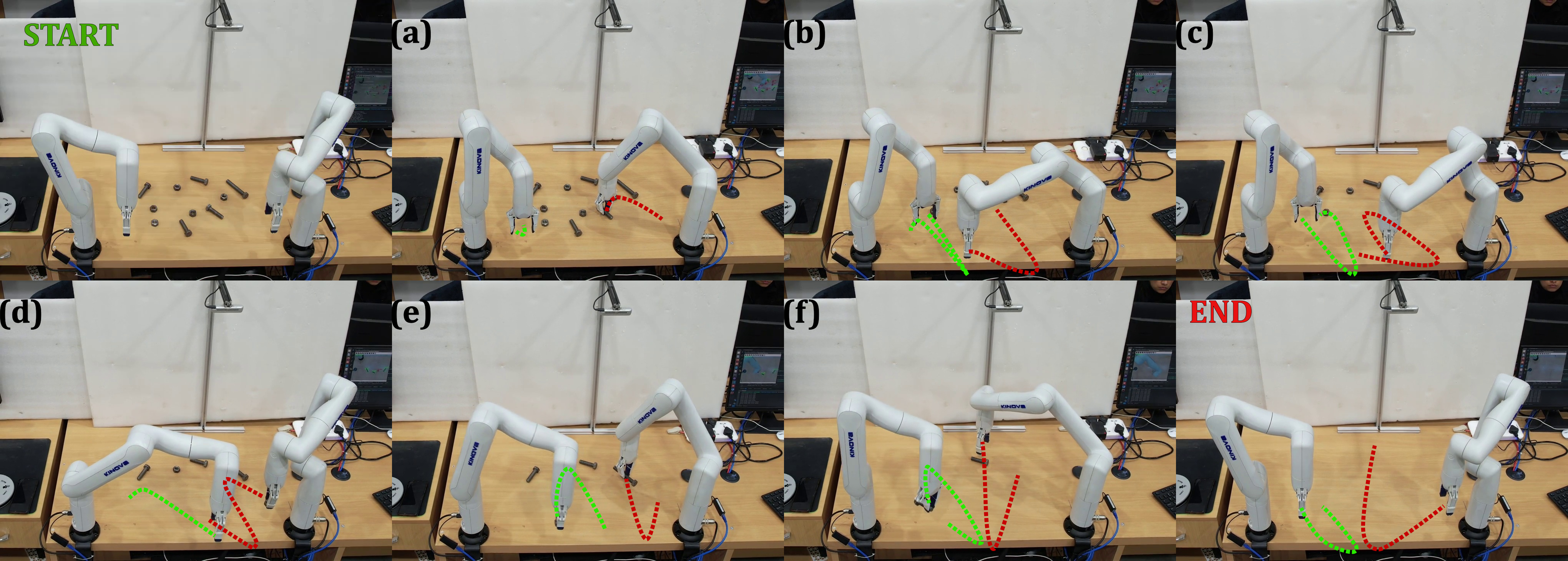}
    \caption{Dual-arm system sorting objects using dynamic task allocation. START: Both arms are at their home positions. (a) Both arms initiate simultaneous manipulation from home positions. (b-c) Coordinated motion planning—left arm (green trajectory) and right arm (red trajectory) execute independent pick-and-place operations. (d-e-f) Continued coordinated execution with iAPF collision avoidance maintaining safe inter-arm distances. END: Return to home positions. The sequence demonstrates zero collision events during compact object manipulation, validating the temporal proposal window mechanism and improved APF framework.}
    \label{Sequential_exp}
\end{figure}
\section{Results and Discussion}
\label{results}
The experimental validation using the hardware as shown in Fig.~\ref{Experimentalsetup} involved comprehensive testing with two 6-DOF Kinova Gen3 Lite manipulators handling industrial fasteners. The performance evaluation of the proposed method on two distinct scenarios. First, homogeneous object allocation where both arms handle identical object types; second, heterogeneous mixed-object scenarios involving simultaneous nuts and bolts processing. The results validate the proposed system's ability to achieve significant coordination improvements over fixed allocation baselines while maintaining zero collision events across all experimental trials, demonstrating practical viability for industrial deployment.

\noindent \textbf{Case I: Homogeneous Object Scenarios - Single Object Type Allocation}

\begin{figure}
    \centering
    \includegraphics[width=0.7\linewidth]{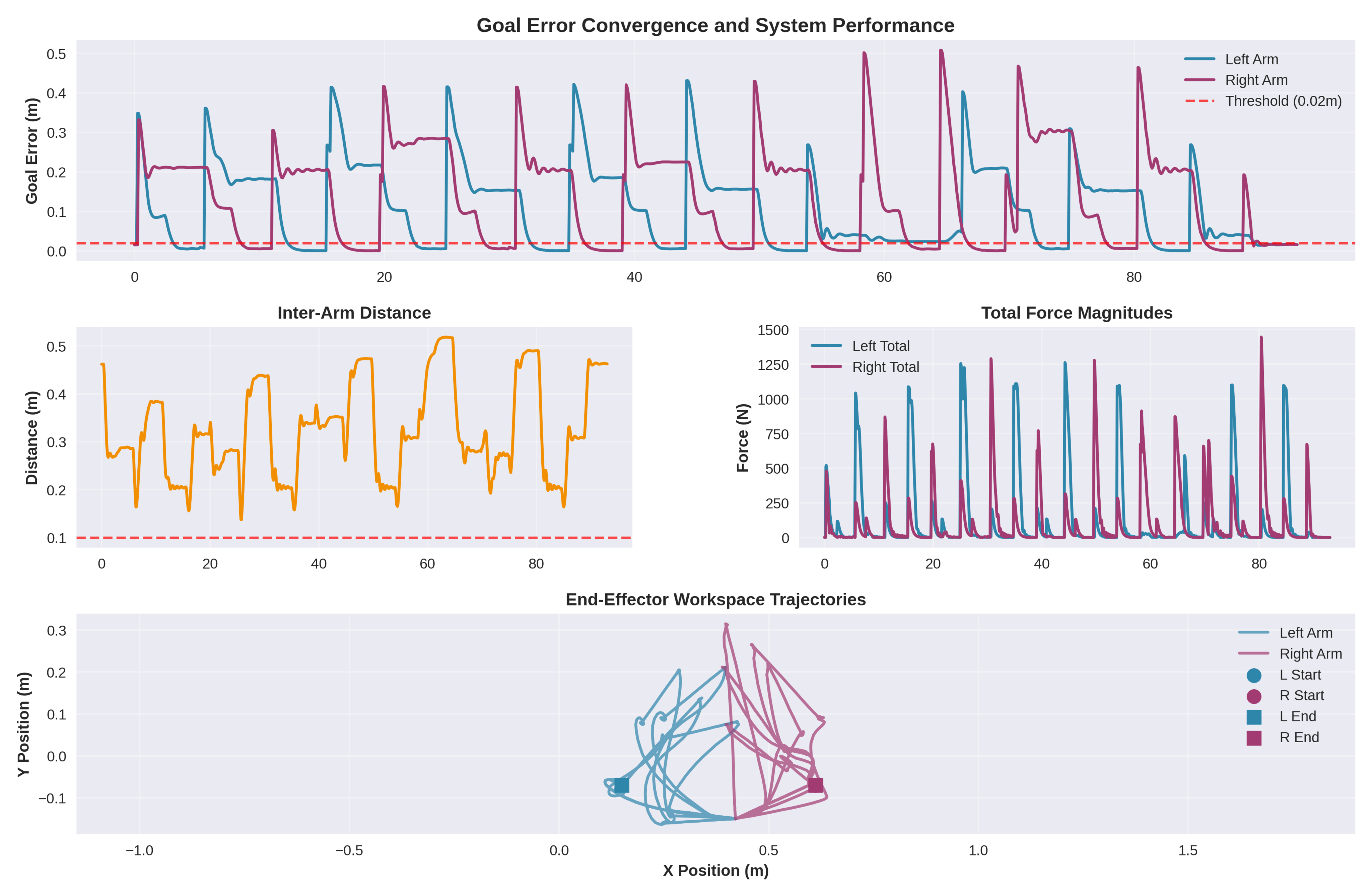}
    \caption{Dual-Arm system Analysis for Dynamic Task Allocation Single Object Type - Homogeneous}
    \label{fig:Combined_Analysis_dynamic_single_object}
\end{figure}

\begin{figure}
    \centering
    \includegraphics[width=0.7\linewidth]{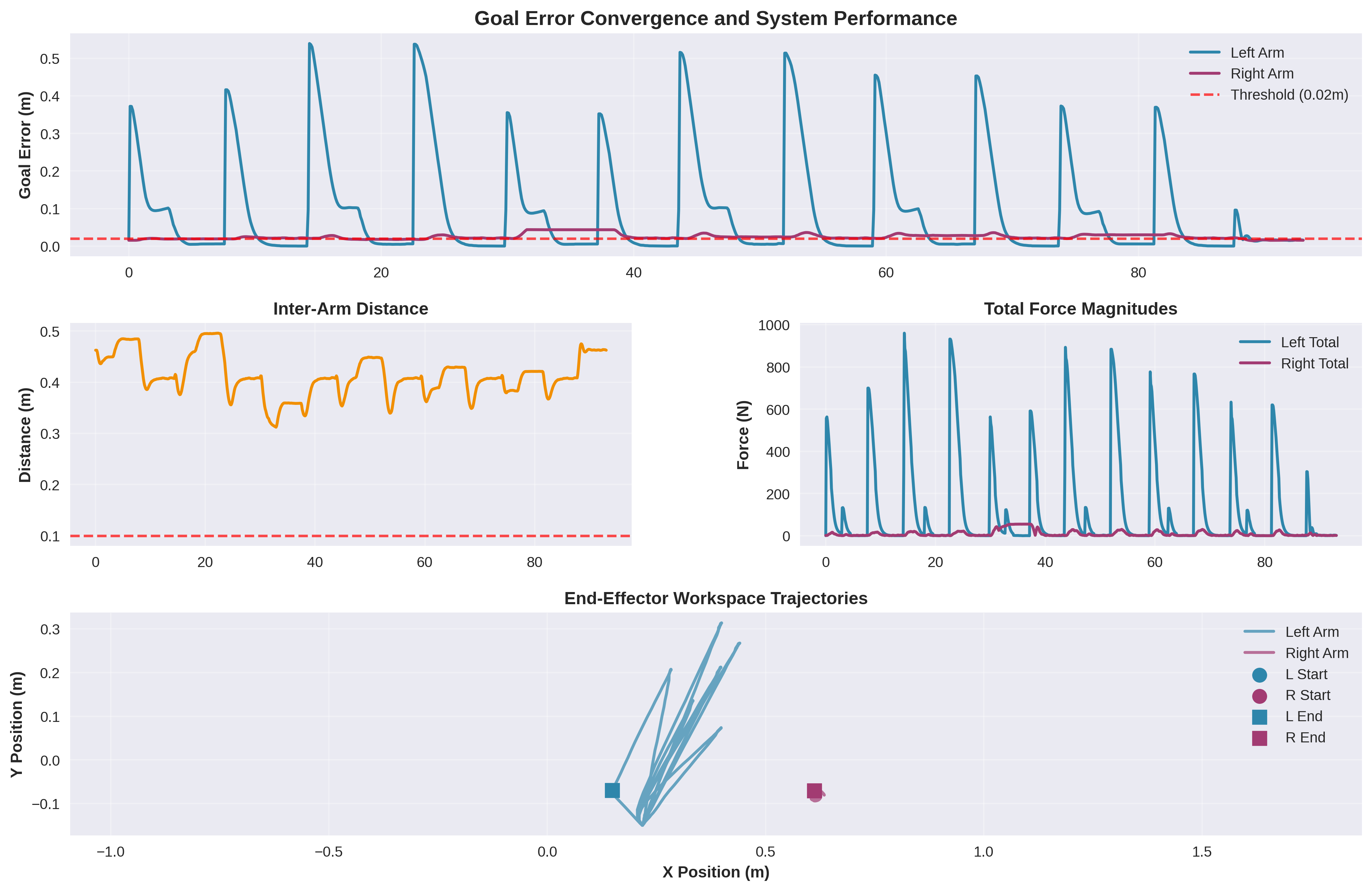}
    \caption{Dual-Arm system Analysis for Fixed Task Allocation Single Object Type - Homogeneous}
    \label{fig:Combined_Analysis_fixed_single_object}
\end{figure}

\noindent The comparative evaluation demonstrates significant performance advantages of the proposed dynamic task allocation system over conventional fixed assignment methods over ten trials, as shown in Table~\ref{tab:performance_comparison_new}. In single object-type scenarios, the dynamic allocation achieves an average efficiency index of $\approx$0.116 m/s compared to $\approx$0.0780 m/s for fixed allocation, representing a $\approx$47\% improvement in task execution rate. This enhancement results from optimal dual-arm coordination, where both manipulators actively contribute to task completion rather leaving one arm idle as in fixed assignment strategies, as shown in Fig.~\ref{fig:Combined_Analysis_dynamic_single_object} and Fig.~\ref{fig:Combined_Analysis_fixed_single_object}.
Proposed method maintains consistent performance across both bolt-only and nut-only (0.116 m/s) scenarios, while effectively utilizing the complete workspace through proximity-based competitive selection with complete dual-arm movement's utilization. 
\begin{table}[h!]
    \centering
    \caption{Performance Comparison: Dynamic vs Fixed Task Allocation.}
    \label{tab:performance_comparison_new}
    \begin{tabular}{lcccc}
    \toprule
    \textbf{Method} & \textbf{Scenario} & \textbf{Total Dist. (m)} & \textbf{Total Time (s)} & \textbf{Avg.Velocity (m/s)}\\
    \midrule\midrule
    \multirow{2}{*}{\textbf{Dynamic}} & Bolt-Only & $7.60\pm0.082$ & $65.134\pm2.01$ & $0.116\pm0.003$ \\
    & Nut-Only & $7.88\pm0.1431$ & $67.87\pm1.06$ & $0.116\pm0.0002$ \\
    \cmidrule{1-5}
    \multirow{2}{*}{\textbf{Fixed}} & Bolt-Only & $6.61\pm0.016$ & $84.65
\pm0.132$ & $0.078\pm0.0001$ \\
    & Nut-Only & $6.42\pm0.0104$ & $82.147\pm0.08$ & $0.0781\pm0.0001$ \\
    \midrule
    \end{tabular}%
    
    \vspace{2pt}
    \footnotesize{\textbf{Note:} Dynamic task allocation achieves $\approx$47\% efficiency improvement through optimal dual-arm coordination. Higher total distance reflects dual-arm coordination covering more workspace versus single-arm operation. (Independent t-test confirms statistical significance (p $<$ 0.001, n=10 trials)).}
\end{table}

In contrast, fixed allocation assigns each arm to specific object types, as shown in Fig.~\ref{fig:Combined_Analysis_fixed_single_object}, resulting in inherent inefficiency $\approx$(0.078 m/s) with only one arm active per object type and average single-arm utilization of $\approx$50\%. These results validate the effectiveness of the temporal proposal window mechanism in achieving superior coordination efficiency for autonomous dual-arm manipulation systems.

\noindent \textbf{Case II : Heterogeneous Object Scenarios - Two Object Type Allocation}

The experimental validation using mixed object scenarios as shown in Fig.~\ref{Sequential_exp} containing both nuts and bolts evaluates real-world industrial applicability. Table.~\ref{tab:heterogeneous_comparison_new}, presents comparative performance metrics between dynamic and baseline methods across ten trials. Each trial include 12 objects(6 nuts, 6 bolts) with compact arrangements.

\begin{table}[h!]
    \centering
    \caption{Mixed Object Scenarios: Hardware Comparison Across Allocation Methods (12 Objects, 10 Trials).}
    \label{tab:heterogeneous_comparison_new}
    \begin{tabular}{lcccc}
    \toprule
    \textbf{Method} & \textbf{Scenario} & \textbf{Total Dist. (m)} & \textbf{Total Time (s)} & \textbf{Avg. Velocity (m/s)} \\
    \midrule\midrule
    \textbf{Dynamic (Proposed)} & Mixed & $15.49 \pm 0.226$ & $135.42 \pm 0.485$ & $0.114 \pm 0.001$ \\
    \textbf{Market-based}       & Mixed & $16.85 \pm 0.315$ & $152.10 \pm 1.450$ & $0.110 \pm 0.002$ \\
    \textbf{Round-robin}        & Mixed & $18.42 \pm 0.412$ & $175.30 \pm 2.100$ & $0.105 \pm 0.003$ \\
    \textbf{Fixed}              & Mixed & $12.57 \pm 0.181$ & $98.40  \pm 0.800$ & $0.128 \pm 0.001$ \\
    \midrule
    \end{tabular}%
    \vspace{2pt}
    \footnotesize{%
    \textbf{Note:} Dynamic, Market-based, and Round-robin achieve $>$89\% coverage with complete dual-arm utilization. 
    One-way ANOVA with post-hoc pairwise t-tests confirms statistical significance across Dynamic, Market-based, and Round-robin methods 
    ($p < 0.001$, $n=10$ trials). Fixed allocation is reported separately given partial task completion ($\approx$55\% coverage.}
\end{table}
\begin{figure}
    \centering
    \includegraphics[width=0.7\linewidth]{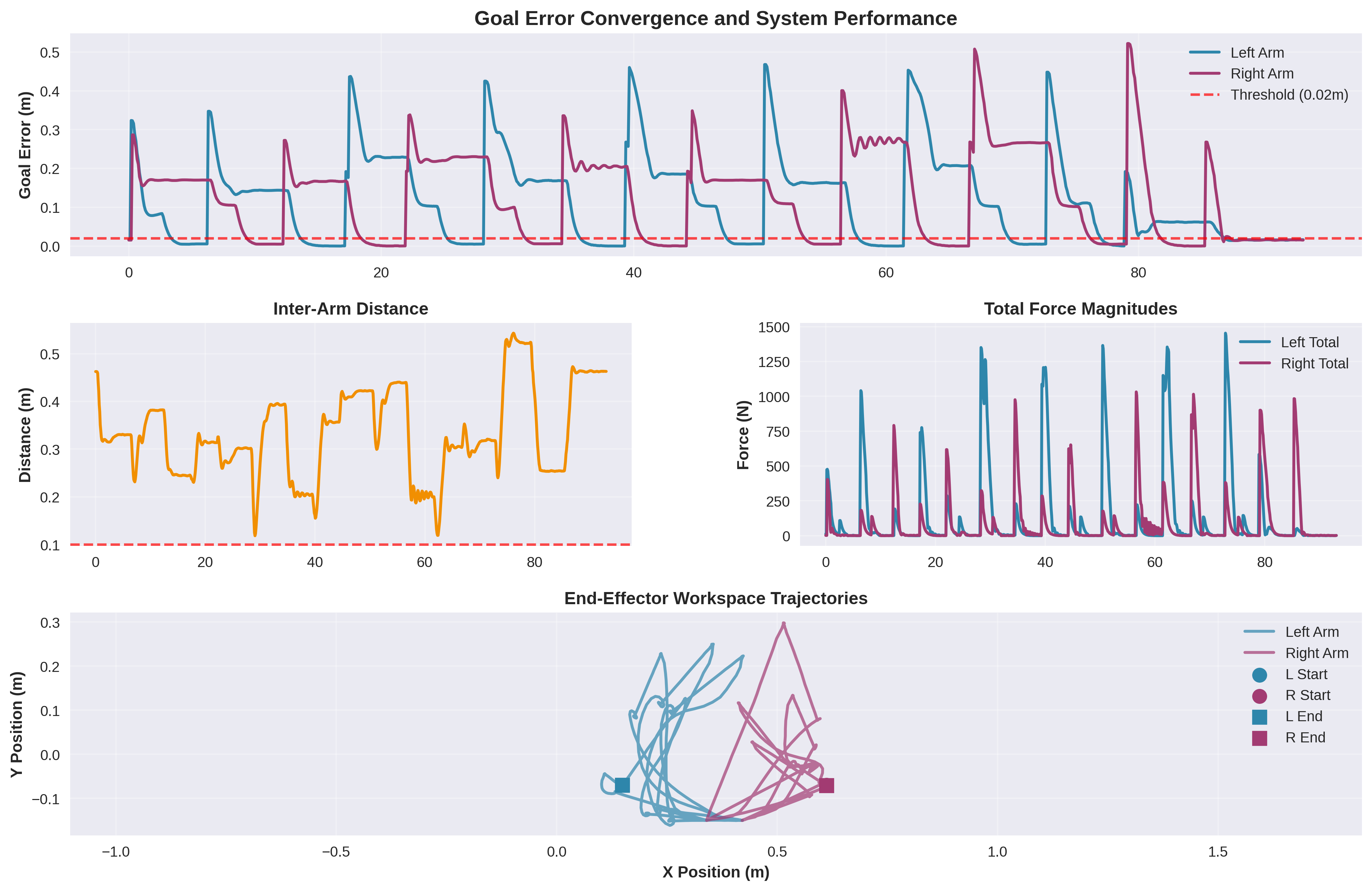}
    \caption{Combined Dual-Arm system Analysis Dynamic Task Allocation}
    \label{fig:Combined_Analysis_dynamic}
\end{figure}

\begin{figure}
    \centering
    \includegraphics[width=0.7\linewidth]{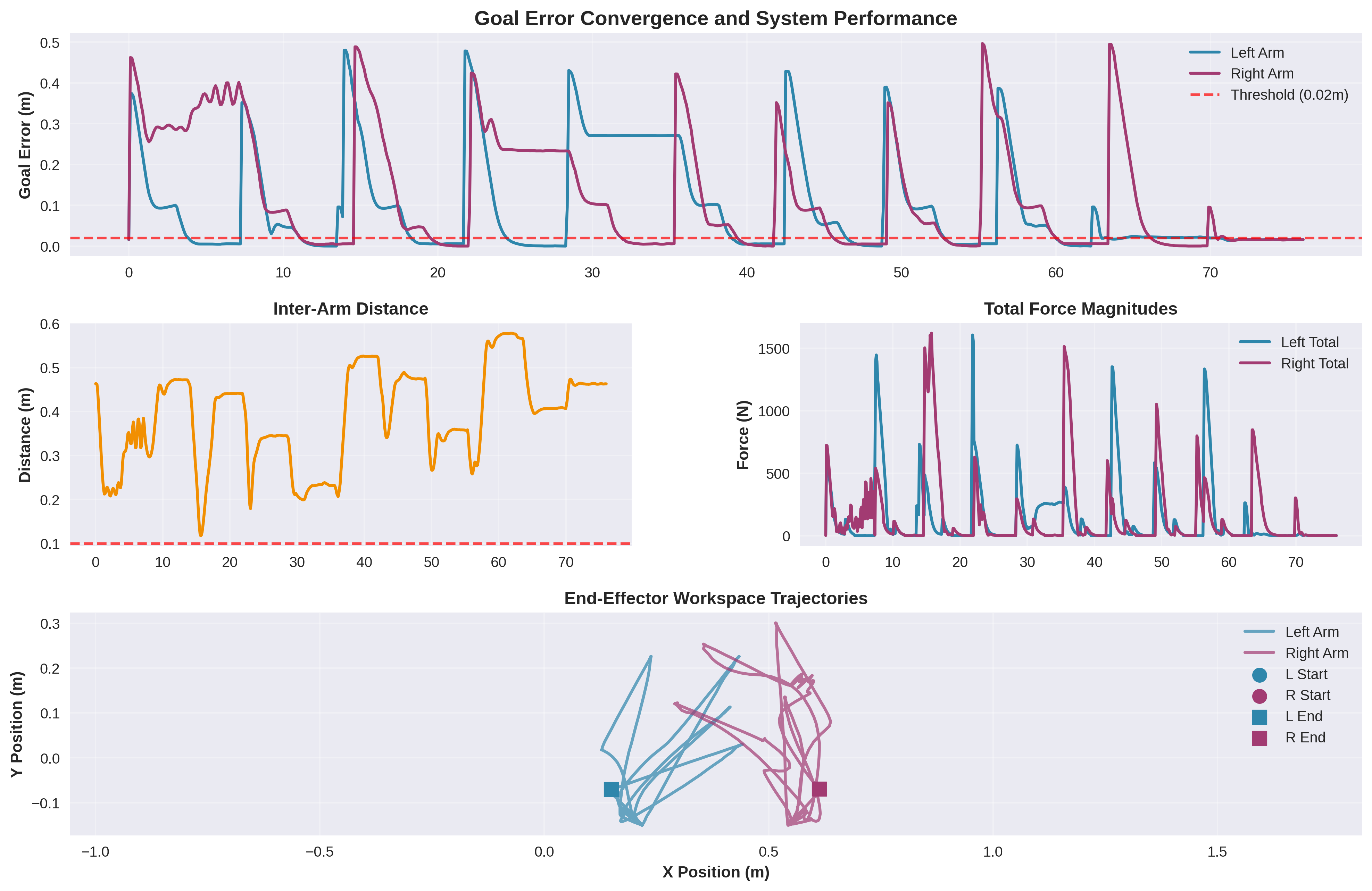}
    \caption{Combined Dual-Arm system Analysis Fixed Task Allocation}
    \label{fig:Combined_Analysis_Fixed}
\end{figure}

Fixed assignment achieved a higher average velocity ($\approx 0.128\,\text{m/s}$) and shorter completion time ($98.40\,\text{s}$), though this is purely an artifact of partial task completion ($\approx 55\%$ coverage) where type-constrained arms operate only within high-density accessible zones, excluding unreachable objects entirely. Among methods achieving complete coverage ($>89\%$), the proposed method ($15.49\,\text{m}$, $135.42\,\text{s}$) outperforms market-based allocation ($16.85\,\text{m}$, $152.10\,\text{s}$) by $8.0\%$ in distance and $11.0\%$ in completion time, and round-robin ($18.42\,\text{m}$, $175.30\,\text{s}$) by $15.9\%$ and $22.7\%$ respectively. These hardware gains are consistent with simulation results, where proposal-based allocation demonstrated superior distance efficiency and $69\%$ lower collision risk over market-based methods. Market-based allocation locks target assignments immediately after each auction without spatial re-evaluation, yielding suboptimal proximity decisions when object distributions shift during execution. Round-robin's alternating assignment entirely disregards spatial relationships, producing the highest travel distance and completion time across all methods. The temporal proposal window introduces modest overhead compared to fixed allocation, but enables universal object handling, superior safety margins, and smoother motion profiles as shown in Fig.~\ref{fig:Combined_Analysis_dynamic}, eliminating oscillatory workspace conflicts observed in fixed assignment as shown in Fig.~\ref{fig:Combined_Analysis_Fixed}. For industrial deployments where reliability and collision safety are paramount, these advantages justify the time overhead.
\section{Conclusion and Future Work}
\label{concl}
This work presented a competitive task allocation system for dual-arm manipulation integrating real-time vision, temporal proposal window management, and improved artificial potential fields for safe industrial coordination. The novel temporal proposal mechanism enables dynamic object-arm pairing through competitive evaluation within five-frame windows, preventing workspace conflicts while optimizing distance-based allocation. Experimental validation with two 6-DOF Kinova Gen3 Lite manipulators demonstrated $\approx$47\% efficiency improvement in homogeneous scenarios and zero collision events across all trials, while mixed object scenarios showed $\approx$11\% efficiency penalty offset by superior motion quality and universal object handling capability.

The temporal proposal window mechanism along with improved APF, successfully established a safety-first approach for manufacturing applications where collision avoidance is paramount, demonstrating practical viability for industrial deployment. Future research will focus on: (1) extending the framework to size-based assembly operations, specifically nut-bolt pairings; this necessitates a transition from the absolute positioning movements in cluttered environments demonstrated here to coordinated, relative end-effector movements for precise threaded fastener insertion and bilateral manipulation. (2) we aim to implement adaptive learning mechanisms for the dynamic adjustment of temporal window parameters and APF force coefficients based on real-time workspace density and task complexity. Finally, (3) we plan to expand our validation to include multi-object assembly sequences with varying geometric complexities to further demonstrate the industrial scalability and generalization capabilities of the proposed system.
\backmatter



\bibliographystyle{sn-mathphys-num}
\bibliography{sn-article}
\end{document}